\documentclass[letterpaper, 10 pt, conference]{ieeeconf}  

\IEEEoverridecommandlockouts                              
\usepackage[utf8]{inputenc}
\usepackage[T1]{fontenc}
\usepackage[disable,textsize=scriptsize]{todonotes}
\usepackage{lipsum}
\usepackage{algorithm}
\usepackage{algpseudocode}
\usepackage{url}
\usepackage{siunitx}

\usepackage{amsmath} 
\usepackage{amsfonts}
\usepackage{amssymb}  

\algblock{ParFor}{EndParFor}
\algnewcommand\algorithmicparfor{\textbf{for}}
\algnewcommand\algorithmicpardo{\textbf{do parallel}}
\algnewcommand\algorithmicendparfor{\textbf{end\ for}}
\algrenewtext{ParFor}[1]{\algorithmicparfor\ #1\ \algorithmicpardo}
\algrenewtext{EndParFor}{\algorithmicendparfor}

\DeclareMathOperator*{\argmax}{arg\,max}

\newcommand\pl[1]{}

\makeatletter
\newcommand\fs@betterruled{%
  \def\@fs@cfont{\bfseries}\let\@fs@capt\floatc@ruled
  \def\@fs@pre{\vspace*{5pt}\hrule height.8pt depth0pt \kern2pt}%
  \def\@fs@post{\kern2pt\hrule\relax}%
  \def\@fs@mid{\kern2pt\hrule\kern2pt}%
  \let\@fs@iftopcapt\iftrue}
\floatstyle{betterruled}
\restylefloat{algorithm}
\makeatother

\title{\LARGE \bf
Learning of Parameters in Behavior Trees for Movement Skills
}

\author{Matthias Mayr$^{1}$, Konstantinos Chatzilygeroudis$^{2,3}$, Faseeh Ahmad$^{1}$, Luigi Nardi$^{1,4}$ and Volker Krueger$^{1}$
	\thanks{$^{1}$
	Department of Computer Science, Faculty of Engineering (LTH), Lund University, SE~221~00 Lund,
		Sweden. E-mail: <firstname>.<lastname>@cs.lth.se.
	}%
	\thanks{$^2$Computer Technology Institute \& Press ``Diophantus'', Patras, Greece.}
	\thanks{$^3$Computer Engineering and Informatics Department (CEID), University of Patras, Greece. E-mail: costashatz@upatras.gr.}
	\thanks{$^{4}$
	Department of Computer Science and Electrical Engineering, Stanford University, CA 94305, USA. E-mail: lnardi@stanford.edu.}
}

\begin{document}

\maketitle
\thispagestyle{empty}
\pagestyle{empty}

\begin{abstract}

Reinforcement Learning (RL) is a powerful mathematical framework that allows robots to learn complex skills by trial-and-error. Despite numerous successes in many applications, RL algorithms still require thousands of trials to converge to high-performing policies, can produce dangerous behaviors while learning, and the optimized policies (usually modeled as neural networks) give almost zero explanation when they fail to perform the task. For these reasons, the adoption of RL in industrial settings is not common.
Behavior Trees (BTs), on the other hand, can provide a policy representation that a) supports modular and composable skills, b) allows for easy interpretation of the robot actions, and c) provides an advantageous low-dimensional parameter space. In this paper, we present a novel algorithm that can learn the parameters of a BT policy in simulation and then generalize to the physical robot without any additional training.
We leverage a physical simulator with a digital twin of our workstation, and optimize the relevant parameters with a black-box optimizer.

We showcase the efficacy of our method with a 7-DOF \emph{KUKA-iiwa} manipulator in a task that includes obstacle avoidance and a contact-rich insertion (peg-in-hole), in which our method outperforms the baselines.

\end{abstract}

\section{Introduction}
Even in well-structured environments, like the ones in industrial settings, a desired feature of a robotic setup is to be able to improve over time and easily adapt to novel situations. Reinforcement Learning (RL)~\cite{chatzilygeroudis2019survey,deisenroth13r} provides the theoretical framework that can allow robots to learn complex skills by trial-and-error. State-of-the-art RL methods, however, require at least a few hundred hours of interaction time to find an effective policy. Moreover, in the early stages of learning, those methods can produce unstable behaviors that can damage the robot or the environment (including human operators). Additionally, the optimized controllers, usually modeled as neural networks, cannot easily be interpreted, and when the robot fails at a task it is difficult to devise mitigation plans. In other words, operators have difficulties understanding \emph{when} a robot is doing \emph{what} and \emph{why} and what kind of reliability can be expected. All things considered, controllers learned via pure RL do not satisfy the requirements for safety and production quality.

\begin{figure}[tpb]
	{
	\setlength{\fboxrule}{0pt}
	\framebox{\parbox{3in}{
		\includegraphics[width=0.95\columnwidth]{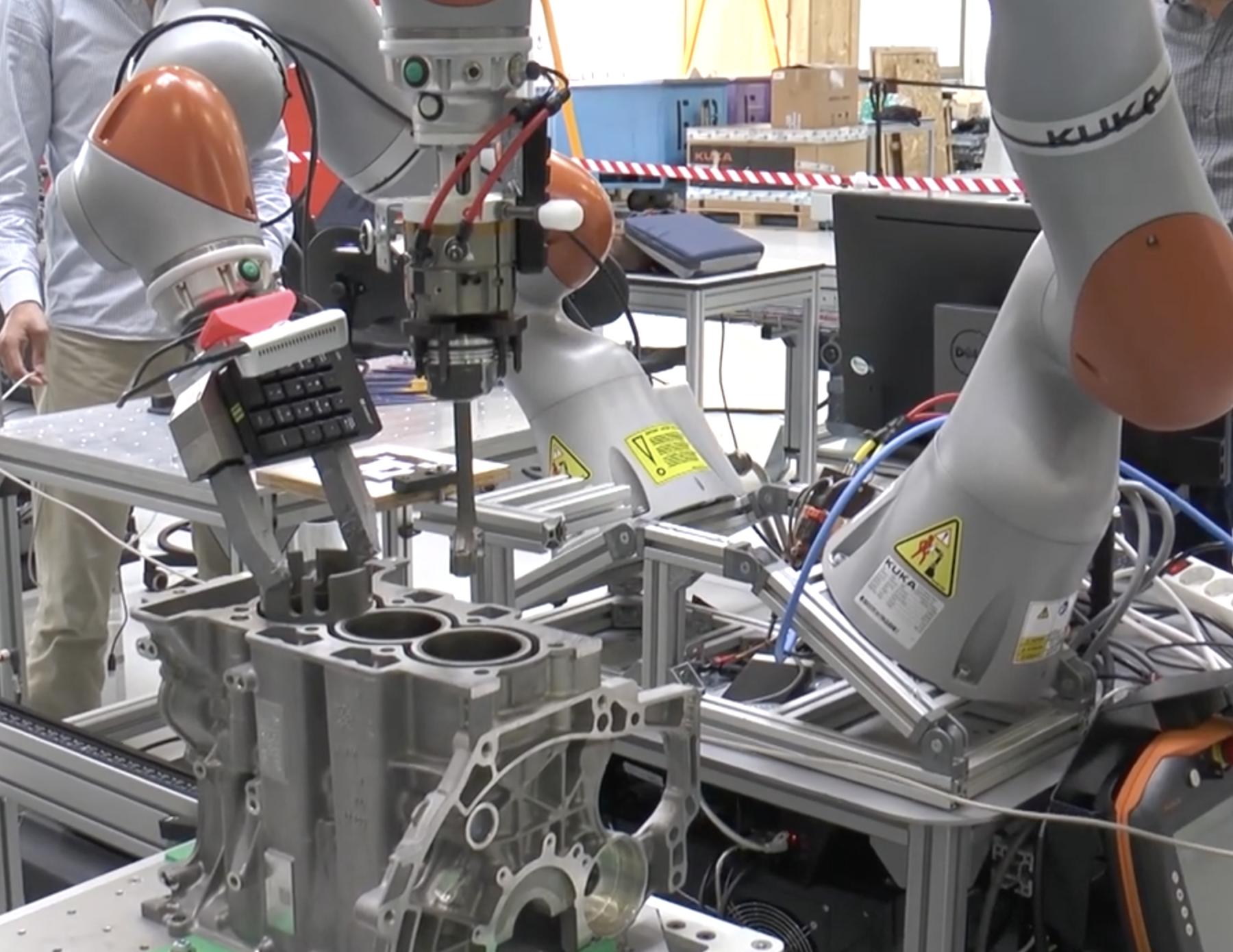}
	}}}
	\caption{The robot setup used for the experiments for engine assembly at PSA: two \emph{KUKA-iiwa} arms mounted on a torso to emulate a configuration similar to a human upper body. Policies are based on BTs and parametric movement skills.}
	\label{fig:robot_setup}
\end{figure}

There exist three main axes for improving the data-efficiency of RL algorithms~\cite{chatzilygeroudis2019survey}: 1) inserting prior knowledge in the policy structure or parameters, 2) learning models of the objective function, and 3) learning or using models of dynamics. Using prior knowledge in the policy can be beneficial as it can a) create an easy to optimize optimization landscape, b) reduce the dimensionality of the search space, or c) provide a strong initial configuration that is close to the optimal value. Learning models using the gathered experience data or using dynamics models (e.g. simulators) allows the agents to perform interactions with the models, and thus reduce the interaction time needed on the physical system~\cite{deisenroth13r,chatzilygeroudis2019survey}.

Many types of policy structures have been proposed in the robot learning literature~\cite{chatzilygeroudis2019survey}. Among them, trajectory-based policy structures~\cite{khansari2011learning,ude2010task,stulp2013robot,ijspeert2013dynamical} have received the most attention. One approach to encode trajectories is to define the policy as a sequence of way-points, while the most widely used approach is using policies based on dynamical systems. The latter type of policies has been used more extensively as they combine the generality of function approximators with the advantages of dynamical systems, such as robustness toward perturbations and stability guarantees~\cite{khansari2011learning,krueger12icra,ude2010task,stulp2013robot}, which are desirable properties of a robotic system. Parametric movement skills (MS) or motion generators~\cite{rovida182iicirsi} are another type of trajectory-based policy structure that explicitly operates in the end-effector space and allows for easy external modulation of their parameters. The advantage of this type of policy representation is that it is modular and easily interpretable. To the best of our knowledge, no prior work is available that combines them with learning.

\pagebreak
In this paper, we propose a novel pipeline that is able to learn interpretable policies with no interaction on the physical robot when a digital twin (DT) of the workspace is available. 
In particular, we propose to use a parametric policy representation that is combining behavior trees (BT)~\cite{colledanchise142iicirsa,colledanchise17ac} with parametric movement skills (MS)~\cite{rovida182iicirsi}, and to learn the parameters of the BT and the MS through reinforcement learning. We call this BT/MS policy representation approach \textit{BTMS}. Since this BTMS policy representation is not differentiable, we solve the policy search using black-box optimization~\cite{chatzilygeroudis172iicirsi,chatzilygeroudis2019survey}. Black-box optimizers are able to optimize based solely on the input and the output and they do not make any assumptions on how the policy is modeled. In addition, we utilize domain randomization techniques~\cite{tobin17iros,mehta20pmlr} in order to generalize the behaviors from the DT to the physical system.

Our formulation has several advantages over prior work, namely:
\begin{itemize}
	\item It can be easily interpreted, and it allows not only to better predict the final robot behavior, but also to efficiently define priors for both structure and parameters for a given task~\cite{rovida172iicirsi};
	\item It allows to decouple a desired policy into sub-policies that can be optimized separately;
	\item It provides a low-dimensional search space, and thus learning on the physical system (either from scratch or fine-tuning) is feasible;
	\item We do \emph{not} require the physical setup to align perfectly with its digital twin.
\end{itemize}

To validate our proposed pipeline, we create the following scenario: a \textit{KUKA-iiwa} 7-degree-of-freedom (DOF) industrial manipulator needs to insert a piston inside an engine while avoiding the obstacles in the workspace (Fig.~\ref{fig:robot_setup}). We emulate this scenario by requiring the manipulator to avoid a big obstacle (the engine block in Fig.~\ref{fig:robot_setup}) in the middle of the workspace, and to insert a peg into a hole with a $5\,mm$ clearance (i.e., difference between the hole size and the peg size) (Fig.~\ref{fig:experiment}).

Our results showcase that the proposed method can learn policies completely in simulation using the DT, and transfer successfully to the physical setup. The results also show that our BTMS policy representation leads to faster learning and better convergence than black-box policies (e.g., neural networks). Finally, using the proposed method we can split the overall policy into sub-policies, learn them separately, and easily combine them to perform the whole task; we show that we can even combine sub-policies where one of them is learned in simulation and the other one on the physical setup.

\section{Related Work}
\label{sec:related_work}

\subsection{Policy Search and Representation}
Model-free policy search approaches have been used in many successful applications in robotics~\cite{deisenroth13r}. They can easily be applied to RL problems with high-dimensional continuous state spaces. However, they are not as sample-efficient and often need hundreds or thousands of episodes~\cite{chatzilygeroudis2019survey, polydoros17jirs, deisenroth13r}.

Model-based methods utilize an internal model of the transition dynamics that allow to optimize a policy without interaction with the physical system. This is generally more efficient~\cite{deisenroth13r, polydoros17jirs}, which is an important criterion since many scenarios in robotics need supervision by a human operator.

The commonly used policy representations include radial basis function networks~\cite{deisenroth13r}, dynamical movement primitives~\cite{ijspeert2013dynamical, ude2010task} and feed-forward neural networks~\cite{deisenroth13r,chatzilygeroudis172iicirsi}. In recent years deep neural nets seem to become a popular policy.
All of them have in common that their final representation can be difficult to interpret. Even if a policy sets a target pose for the robot to reach, it can be problematic to know how it reacts in all parts of the state space.

\subsection{Behavior Trees in Manipulation}

Behavior trees are a policy representation that can be used for both planning and execution. They have been first described in computer games~\cite{colledanchise142iicirsa,iovino2020survey} and have successful applications in robotics~\cite{rovida172iicirsi,rovida182iicirsi,marzinotto142iicrai,7989070,iovino2020survey}.

They are often chosen in skill-based systems~\cite{rovida172iicirsi} because they are modular, interpretable and reactive. However, both their structure and parameters are often hard-coded for a specific task~\cite{rovida182iicirsi,marzinotto142iicrai,colledanchise142iicirsa,7989070,iovino2020survey}. Frameworks like \textit{extended BTs}~\cite{rovida172iicirsi} allow to find both through planning. But this is limited to the capabilities of the reasoning system and can be challenging for contact-rich tasks. Furthermore, there is no mechanism to improve the task performance~\cite{rovida172iicirsi}.

Learning BTs is mostly exercised in game AI. A recent survey of BTs in robotics and AI~\cite{iovino2020survey} provides a detailed overview of the BT learning methods.
In short, existing approaches mostly focus on evolving the structure of a tree~\cite{lim10aec, perez11aeca, colledanchise18itg} or learning the behavior of \emph{control flow nodes}~\cite{fu16a}. 
However in this work we are assuming that the structure is given \textit{a priori}, e.g., through a framework like~\cite{rovida172iicirsi}, to assure better predictability, and learn the parameters of the BT through trial-and-error.

To the best of our knowledge, there is no previous work of applying reinforcement learning with BT policies in the context of manipulation~\cite{iovino2020survey}.

\section{Approach}
\label{sec:approach}
Overall, we propose to combine BTs with parametric MS to form a novel policy representation to be learned via reinforcement learning. Most of the learning happens in simulation, and we use domain randomization techniques so that the learned controllers generalize to the physical setup.

In the following subsections, we lay out our choices and approach. In Sec.~\ref{sec:bg.rc} we present the Cartesian controller of the end-effector of our manipulator. In Sec.~\ref{sec:motion-generator} we describe the parametric MS formulation used in our method that computes commands for the low-level Cartesian controller (Sec.~\ref{sec:bg.rc}). In Sec.~\ref{sec:behavior-trees} we describe the behavior tree formulation and how we use it in our setup to incorporate the parametric MS. Finally, in Sections~\ref{sec:pf}, and~\ref{sec:bg.dr} we describe the policy search formulation, and the domain randomization techniques used for our approach. The whole procedure is shown in Algorithm~\ref{alg:param-opt}.

\begin{algorithm}[tb]
	{
		\setlength{\fboxrule}{0pt}
		\framebox{\parbox{3in}{
				\caption{\strut BT Learning Process}
				\begin{algorithmic}[1]
					\Procedure{BTMS Policy Learning}{}
					\State Define policy $\pi : \mathbf{x} \times \mathbf{\theta} \rightarrow \mathbf{u}$
					\For{$i \leq it_{max}$}\Comment{$\argmax_\mathbf{\theta} \mathbb{E}  \big[ J(\mathbf{\theta}) \big]$}
					\State $\mathbf{\theta}_i \sim \mathcal{N}(\mathbf{m}_k, \sigma_k^2\mathbf{C}_k)$ \Comment{CMA-ES} \label{alg:param-opt:sample}
					\State $R_{i} = \varnothing$
					\ParFor{$e \leq evals$}\Comment{Episodes}					\label{alg:param-opt:dr-loop}
						\State Set random initial robot state $\mathbf{x}_0$	\label{alg:param-opt:init-robot}
						\State Set random initial simulation state				\label{alg:param-opt:init-env}
						\State $D_{i,e} = \varnothing$
						\For{$j = 0 \rightarrow T-1$}
						\State $\mathbf{u}_j = \pi(\mathbf{x}_j|\mathbf{\theta}_i)$
						\State $\mathbf{x}_{j+1} = \text{execute\textunderscore on\textunderscore robot} (\mathbf{u}_{j})$
						\State $D_{i,e}=D_{i,e} \cup\left\{(\mathbf{x}_j, \mathbf{u}_j)\right\}$
						\EndFor
					\State $R_{i} = R_{i} \cup \text{reward}(D_{i,e})$\Comment{Sec.~\ref{sec:reward}}
					\EndParFor
					\State $r_i = mean(R_i)$	\label{alg:param-opt:r_mean}
					\EndFor
					\State $\theta^* = \text{get\textunderscore mean\textunderscore of\textunderscore last\textunderscore population}()$	\label{alg:param-opt:pop_mean}
					\EndProcedure
				\end{algorithmic}
				\label{alg:param-opt}
}}}
\end{algorithm}

\subsection{Robot Control}
\label{sec:bg.rc}
For our arm motions, we are controlling the end effector in Cartesian space. The desired behavior is defined in task coordinates \(\mathbf{x}_{ee}\). A given joint configuration of the robot \(\mathbf{q}\) can be converted into a Cartesian pose using forward kinematics \({\mathbf{x}_{ee} = \mathbf{f}_{fk}(\mathbf{q})}\). This allows to calculate the pose error \({\tilde{\mathbf{x}} 
}\) between \(\mathbf{x}_{ee}\) and a time-varying \emph{virtual equilibrium} or \emph{attractor point} \(\mathbf{x_d}\).

With inspiration from Cartesian impedance control, we formulate the joint commands given joint positions \(\mathbf{q}\) and joint velocities \(\dot{\mathbf{q}}\) according to
\begin{equation}
    \label{formula:control}
    \dot{\mathbf{q}_c}=\mathbf{J}(\mathbf{q})^{T} \left(\mathbf{K}_{d} \tilde{\mathbf{x}}+\mathbf{D}_{d} \mathbf{J}(\mathbf{q}) \dot{\mathbf{q}}\right),
\end{equation}
with \(\mathbf{J}(\mathbf{q})\) being the Jacobian for the end effector in the absolute inertial coordinate frame, \(\mathbf{K}_{d}\) the stiffness matrix and \(\mathbf{D}_{d}\) the damping matrix.

Furthermore, we require the real robot to adhere to safety requirements for safe collaboration with humans at the workstation. This includes an adjustable maximum joint velocity and limiting the torques that can be exerted by the robot arm. To achieve that, we use a joint impedance controller on the real robot. This also means that robot motions can not be executed as accurately as with non-collaborative position-controlled robots. Due to the lack of a joint velocity interface, we integrate the joint velocity commands \(\dot{\mathbf{q}_c}\) to obtain joint position commands \(\mathbf{q}_c\).

In order to have an accurate representation in simulation, the robot model has strict limits on the joint torques.

\subsection{Parametric Movement Skills}
\label{sec:motion-generator}

The reactive adaptation of movement skills in combination with BTs can be used to allow more robust assembly task automation~\cite{rovida182iicirsi}. Rovida et al. advocate to control the end effector in Cartesian space.

A Cartesian goal point \(\mathbf{x}_g\) can be set and a path between the current end effector position \(\mathbf{x}_{ee}\) and the goal position \(\mathbf{x}_g\) is calculated. The attractor point \(\mathbf{x}_d\) is moved along this path with a path velocity \(v_p\). Furthermore, the pose \(\mathbf{x}_d\) can be overlayed with additional motions that can be configured by the BT. We use an Archimedes spiral, which is a common method in peg-in-hole assembly tasks. The formulation follows~\cite{park17itie} and defines a spiral trajectory in polar coordinates
\begin{equation}
\label{formula:spiral}
    \left\{\begin{array}{l}
        \delta \alpha=\frac{v_s}{r_{i}} \\
        r_{i+1}=r_{i}+\delta \alpha \frac{c}{2 \pi},
    \end{array}\right.
\end{equation}
with a path velocity \(v_s\) and a pitch, i.e. the distance between lines in the spiral, \(c\).

One major advantage of this approach is that the movements can be defined in the task frame~\cite{rovida182iicirsi}.
This allows to easily relocate a task frame through planning or sensing and still have a valid motion configuration.
Furthermore, it offers an easier transfer of skills to other robots~\cite{topp182iicirsi}.
We use motion generators as the movement skills in our policy representation, please refer to~\cite{rovida182iicirsi} for details.

\subsection{Behavior Trees}
\label{sec:behavior-trees}
A Behavior Tree (BT)~\cite{colledanchise17ac} is a plan representation and execution tool.
A BT can be defined as a directed acyclic graph $G(V,E)$ with $|V|$ nodes and $|E|$ edges. It consists of \emph{control flow nodes} or \emph{processors}, and \emph{execution nodes}.
The classical formulation defines four types of \emph{control flow nodes}: 1)~\emph{sequence},  2)~\emph{selector}, 3)~\emph{parallel} and 4)~\emph{decorator}~\cite{marzinotto142iicrai}.
A BT has always an initial node with no parents, defined as \emph{Root}, and one or more nodes with no children, called \emph{leaves}.

The execution of a BT is done by periodically injecting a \emph{tick} signal into the \emph{Root}. The signal is routed according to the \emph{control flow nodes} and the return statements of the children. By convention, the signal propagation goes from left to right. This can be exploited to set priorities.

A \emph{sequence} node can be seen as a logical \emph{AND}: it succeeds if all children succeed, and fails if one child fails. The \emph{selector}, also called \emph{fallback} node, represents a logical \emph{OR}: It fails only if all children fail. If one child succeeds, the remaining ones will not be ticked. The \emph{parallel} control flow node forwards ticks to all children and fails if one fails. A \emph{decorator} allows to define custom functions. Leaves of the BT are the \emph{execution nodes} that, when ticked, perform an operation and output one of the three signals: \emph{success}, \emph{failure} or \emph{running}. In particular, execution nodes subdivide into 1)~\emph{action} and 2)~\emph{condition} nodes. An action performs its operation iteratively at every tick, returning \emph{running} while it is not done, and \emph{success} or \emph{failure} otherwise. A condition never returns running: it performs an instantaneous operation and returns always \emph{success} or \emph{failure}.

In this paper, to assure predictability of the robot behaviors, we assume that a structure is given \emph{a-priori}, only the parameters of the BT and the motion skills need to be determined. We utilize nodes and conditions in the BT to set the configuration of the motion generator in a reactive and easy to understand way. Fig~\ref{fig:bt-task} shows the BT that we are using.

\subsection{Policy Optimization}
\label{sec:pf}
In order to optimize for policy parameters, we adopt the policy search formulation~\cite{deisenroth13r,chatzilygeroudis2019survey,chatzilygeroudis172iicirsi}. In general, we model the system as a dynamical system of the form:
\begin{equation}
\label{formula:dynamics}
\mathbf{x}_{t+1}=\mathbf{x}_{t}+M(\mathbf{x}_{t}, \mathbf{u}_{t})+f(\mathbf{x}_{t}, \mathbf{u}_{t}
),
\end{equation}
with continuous valued states $\mathbf{x} \in \mathbb{R}^E$ and continuous valued actions $\mathbf{u} \in \mathbb{R}^U$. The transition dynamics can generally be modeled as a combination of a simulation of the robot $M(\mathbf{x}_t, \mathbf{u}_t)$ and $f(\mathbf{x}_{t}, \mathbf{u}_{t})$ which models the residual between simulation and reality.
Here, we do not model the function $f$. Instead, we use domain randomization for finding a policy that is robust enough such that we can ignore $f$.

The goal is to find a policy $\pi, \mathbf{u} = \pi(\mathbf{x}|\mathbf{\theta})$ with policy parameters $\mathbf{\theta}$ such that $\mathbf{u}$ maximizes the expected long-term reward when executing the policy for $T$ time steps:
\begin{equation}
\label{formula:long_term_reward}
J =\mathbb{E} \left[\sum_{t=1}^{T} r(\mathbf{x}_{t}) | \mathbf{\theta}\right]
,
\end{equation}
with $r(\mathbf{x}_{t})$ being the immediate reward for being in state $\mathbf{x}$ at time step $t$.


In our approach the movement skills in our policy $\pi$ define the movements of the manipulator in Cartesian space in terms of goal points.
A BT is used to ensure the configuration depending on the state of the robot~\cite{rovida172iicirsi, marzinotto142iicrai}.

Since our policy representation is not differentiable, we frame the optimization of Eq.~\eqref{formula:long_term_reward} as a black-box optimization, and seek the maximization of a reward function $J(\theta)$ by only using measurements of the function.
Covariance Matrix Adaptation Evolution Strategy (CMA-ES)~\cite{hansen2006towards} is a stochastic, derivative-free method for numerical optimization of non-linear or non-convex continuous optimization problems.
It models a population of points as a multivariate normal distribution.
CMA-ES performs three steps at each generation $k$, we defer to~\cite{hansen2006towards} for details:
\begin{enumerate}
    \item Reproduction: sample $\lambda$ new offspring according to a multi-variate Gaussian distribution of mean $\mathbf{m}_k$ and covariance $\sigma_k^2\mathbf{C}_k$, that is,    $\mathbf{\theta}_i \sim \mathcal{N}(\mathbf{m}_k, \sigma_k^2\mathbf{C}_k)$ for     $ i \in 1, ..., \lambda$;
    \item Truncation selection: rank the $\lambda$ sampled candidates based on their performance $J(\theta_i)$ and select the fittest $\mu$ individuals with $\mu \le \lambda$;
    \item Gaussian update: to reflect the distribution of the $\mu$ best candidates, compute $\mathbf{m}_{k+1}$ by averaging the $\mu$ individuals: $\mathbf{m}_{k+1} = \frac{1}{\mu} \sum_{i=1}^{\mu} \mathbf{\theta}_i$, and $\sigma_{k+1}^2\mathbf{C}_{k+1}$.
\end{enumerate}
In our experiments, we use BIPOP-CMA-ES with restarts~\cite{hansen2009benchmarking,cully18j}\footnote{We use the open source CMA-ES integration in the Limbo package~\cite{cully18j}.}, which is a refinement over canonical CMA-ES.
At convergence, it restarts the optimization process
with a bigger population $\lambda$ to increase exploration. In our implementation, we sample a new parameter configuration $\theta_i$ at the beginning of an iteration in line~\ref{alg:param-opt:sample} of Algorithm~\ref{alg:param-opt}. When all $\lambda$ offspring of a population have been evaluated, the algorithm performs the truncation selection and Gaussian update.

Examples of parameters to optimize for are location of goal points in Cartesian space or thresholds in conditions in the BT.

\subsection{Domain Randomization}
\label{sec:bg.dr}
Domain randomization is used to bridge the \emph{reality gap} between the digital twin and the physical robot.
The idea of domain randomization is to introduce enough variability into the simulation such that the real physical robot may appear as just another variation of the simulation \cite{tobin17iros,nguyen18iros,chen18iros}. It can also avoid overfitting to a specific solution that might perform very well in simulation, but would fail on the real robot~\cite{koos2012transferability}.
Furthermore, we are also interested to learn robust policies that can address uncertainties in the environment.
In our scenarios, a given policy is evaluated in multiple simulations. See line~\ref{alg:param-opt:dr-loop} in Algorithm~\ref{alg:param-opt}. At the beginning of each evaluation, a varying starting position for the robot arm (line~\ref{alg:param-opt:init-robot}) and a random displacement of objects in the workplace (line~\ref{alg:param-opt:init-env}) is applied. We calculate the mean reward of all simulations in line~\ref{alg:param-opt:r_mean} and assign it as the value of the objective function.
Furthermore, at the end of a learning procedure in line~\ref{alg:param-opt:pop_mean}, we get the mean of the last population of CMA-ES rather than the best performing offspring. The reasoning is that it should be a more robust configuration and the chances that it succeeds on the physical system are higher.
This allows for our policy to generalize to inevitable differences between the physical setup and its DT.

\section{Experimental Results}
\label{sec:experiments}
We evaluate our approach with a \emph{KUKA-iiwa} 7-DOF manipulator (see Fig.~\ref{fig:robot_setup}).
The robot is controlled by our own re-implementation of the motion generator described in Sec.~\ref{sec:motion-generator}.
We utilize the \emph{DART} simulator~\cite{lee18joss}.

\begin{figure*}[tpb]
	\centering
	{
		\setlength{\fboxrule}{0pt}
		\framebox{\parbox{0.98\textwidth}{
		\centering
		\includegraphics[width=0.95\textwidth]{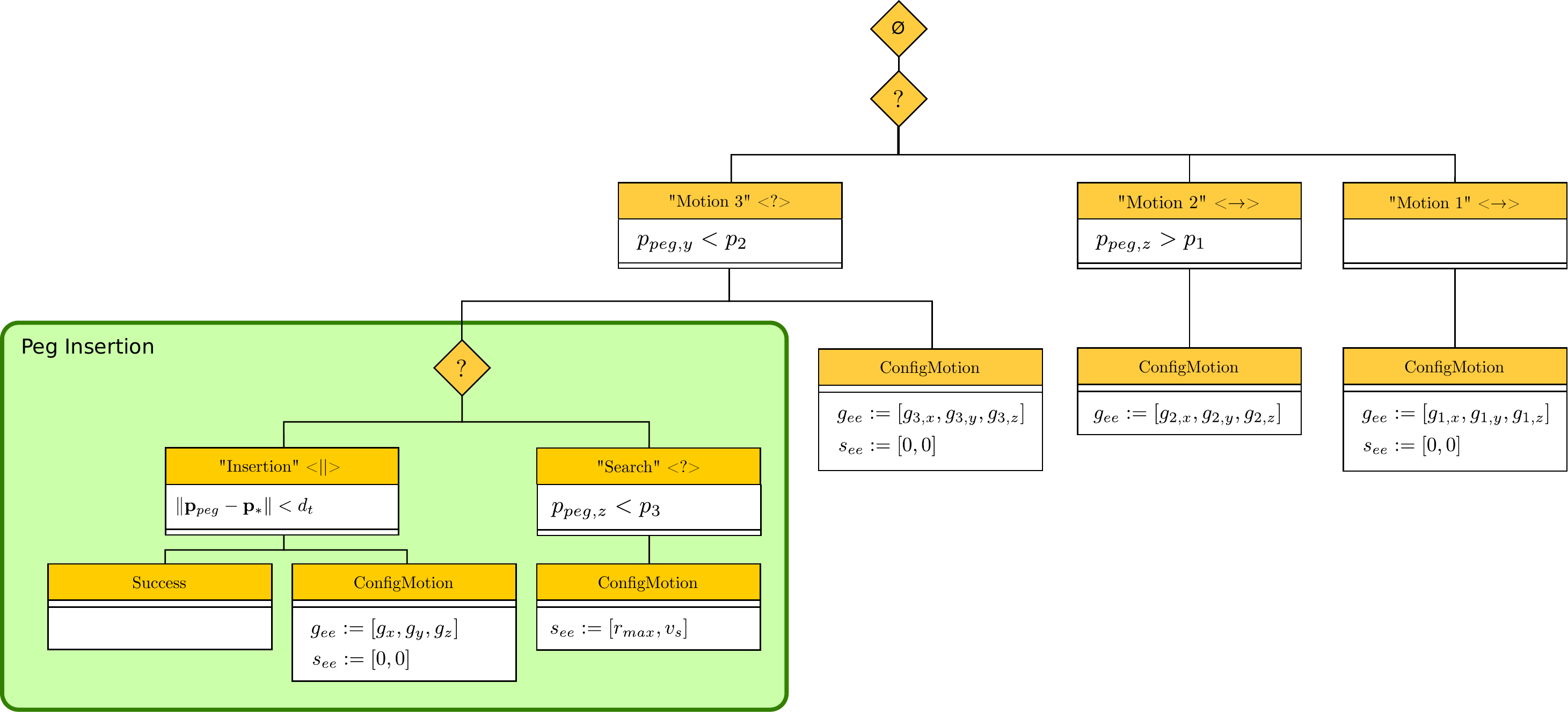}
		\caption{The behavior tree used for the combined task in the extended BT format~\cite{rovida172iicirsi}. Each node has conditions shown in the upper half and effects shown in the lower half. The two explicit \emph{control flow nodes} are \emph{selectors} indicated by a question mark. They forward a tick to their children from the left to the right until one returns running or success. Hence "Motion 1" is only executed if the conditions of the other children do not hold. The top right part of the tree handles the obstacle avoidance. The bottom left part of the tree handles the peg-in-hole task.}
		\label{fig:bt-task}
	}}}

\end{figure*}

For our experiments we use a mock-up of our workstation shown in Fig.~\ref{fig:experiment}.
The goal is to reach a target position (\(g_3\)) on the other side of an obstacle (e.g., the engine block in Fig.~\ref{fig:robot_setup}) and perform a peg insertion on that side.
In order for the task to succeed, two parameters in the BT and several movement skill parameters need to be learned.

The BT is shown in Fig.~\ref{fig:bt-task}. It consists of the bottom left part that models the peg insertion marked in green and the top right part that is parameterized to avoid the obstacle.
Learning the whole task with a real robot arm would include many undesired collisions with the environment.
Instead, we use the property of BTs that a node in the tree can be replaced with any number of nodes~\cite{colledanchise142iicirsa}.
When learning the obstacle avoidance task, the bottom-left BT for the peg insertion is replaced with a condition that reports \emph{success} when the end effector is close to the hole.
This way, we can separate the two tasks.
Therefore, we can learn the obstacle avoidance task in simulation and minimize the usage of the real system and avoid the risk of damages.
We then learn the peg insertion in simulation as well as with the real robot and compare the results.
Here, learning the peg insertion does not pose a danger for the environment or the robot.
Finally, we re-combine the two separately learned policies two evaluate the whole task.

\subsection{Rewards}
\label{sec:reward}

For our tasks, we use the following intuitive rewards:
\begin{enumerate}
	\item Finish the task;
	\item Closeness to the goal;
	\item Negative reward close to obstacles;
	\item Closeness to the hole of the workpiece.
\end{enumerate}
In principle, these rewards can be provided by the DT.
The DT already provides the information about the goal and if an object is to be avoided.\\
\textbf{1.~Finish the task:} As explained in Section~\ref{sec:behavior-trees}, a BT can succeed, run or fail.
This feedback can be used to easily determine if the task was successfully finished. For that purpose we use a task-dependent fixed reward.\\
\textbf{2.~Goal position:} We use the Euclidean distance between the position of the end effector and the target.
We utilize the same function used in~\cite{deisenroth11p2icicml} and~\cite{chatzilygeroudis172iicirsi}, an exponential function:
\begin{equation}
    \label{formula:reward_exp}
    r_{g}(\mathbf{x})=\exp \left(-\frac{1}{2 \sigma_{c}^{2}}(\left\|\mathbf{p}_{peg,\mathbf{x}}-\mathbf{p}_{*}\right\| + d_g)\right),
\end{equation}
where \(\sigma_{c} = 0.4\), \(\mathbf{p}_{peg,x}\) corresponds to the position of the peg in state \(\mathbf{x}\) and \(\mathbf{p}_{*}\) is the goal position. We add a minimum distance \(d_g = 0.25\) to obtain a steeper slope next to the goal.

\textbf{3.~Collisions:} Keeping a safe distance to a certain object in the environment like the table and the engine in Fig.~\ref{fig:robot_setup} is often a basic requirement when parameterizing a policy.
This can be easily achieved without additional sensing using the simulation.
The objects are known in the DT, and a negative reward close to them can easily be defined:
\begin{equation}
    \label{formula:avoidance}
    r_{a}(\mathbf{x})= - \frac{1}{\left( d(\mathbf{p}_{peg,\mathbf{x}}, \mathbf{p}_{obj}) + d_a \right)^{2}},
\end{equation}
with \(d_a = 0.03\) to avoid a division by zero and calculating the shortest distance \(d(\mathbf{p}_{peg,\mathbf{x}}, \mathbf{p}_{obj})\).
Furthermore, the same negative reward can be utilized to avoid areas that might be dangerous for the robot or an operator.\\
\textbf{4.~Closeness to the hole: } This quite localized reward acts as an additional attractor towards the desired assembly goal location. We use
\begin{equation}
    \label{formula:hole}
    r_{h}(\mathbf{x})= \frac{1}{2\left(d(\mathbf{p}_{peg,\mathbf{x}}, \mathbf{p}_{h}) + d_h\right)},
\end{equation}
with \(d_h = 0.006\) and the shortest distance function \(d(\mathbf{p}_{peg,\mathbf{x}}, \mathbf{p}_{h})\). \(\mathbf{p}_{h}\) is a hyperrectangle in the center of a hole with two sides of \SI{2}{\mm} and stretching from the bottom to \SI{1}{\mm} below the surface.

\subsection{Movements with Avoidance of a Static Obstacle}

\begin{figure}[tpb]
    \centering
	{
	\setlength{\fboxrule}{0pt}
	\framebox{\parbox{3in}{
	\includegraphics[width=0.85\columnwidth]{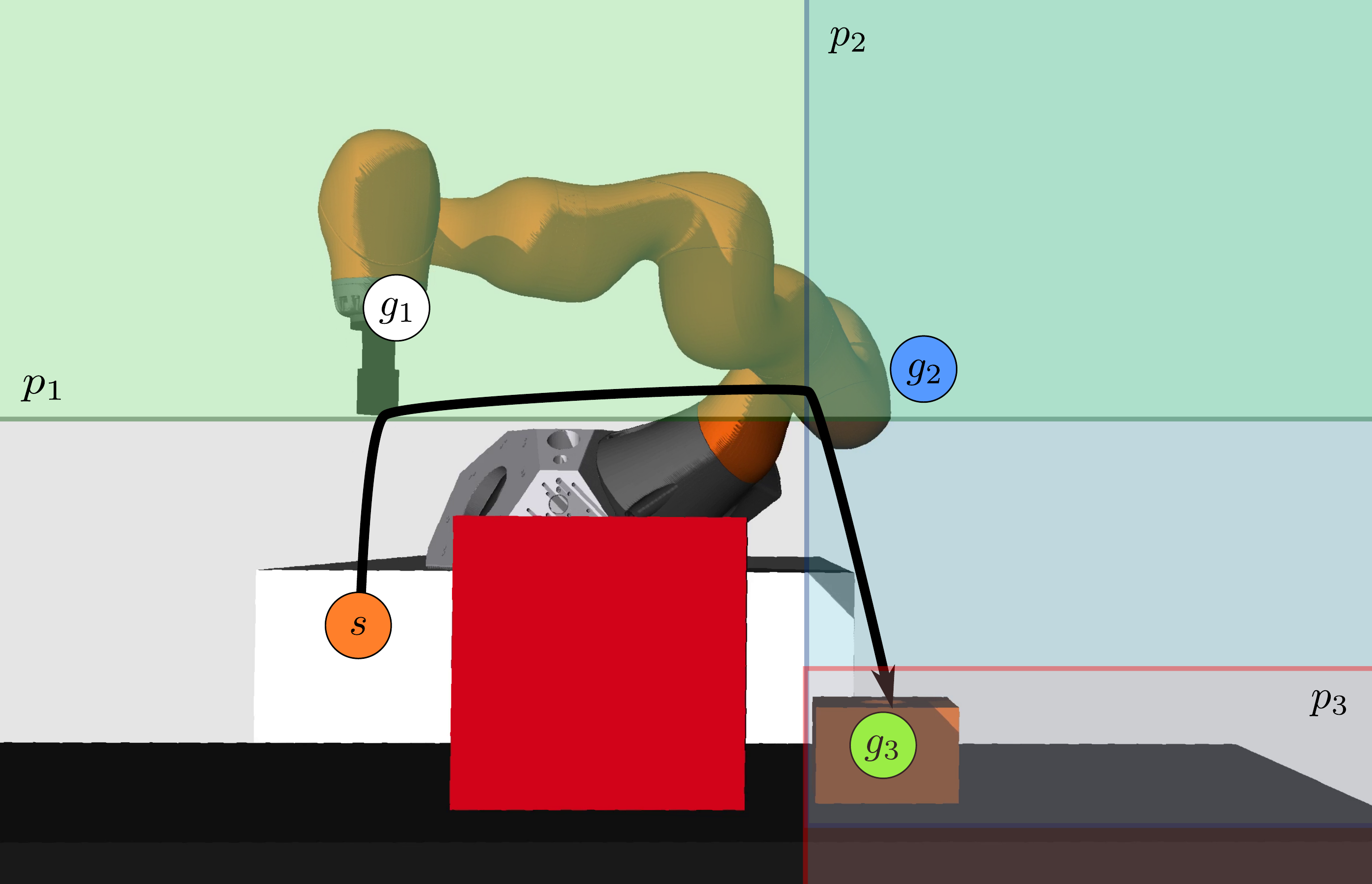}
	\caption{The setup of the experiment in simulation in one possible motion configuration. Two thresholds that are set by the parameters \(p_1\) and \(p_2\) are visualized, as well as the parametric goal points \((g_1, g_2)\). The trajectory between the start point \(s\) and the target \(g_3\) is outlined.}
	\label{fig:experiment}
	}}}
\end{figure}

Intuitively, the shape of the desired trajectory suggests to use two intermediate goal points and two conditions for nodes in the BT.
The pre-defined structure of the BT can be seen in Fig.~\ref{fig:bt-task}, top-right, and an interpretation is shown in Fig.~\ref{fig:experiment}. In the configuration shown in Fig.~\ref{fig:experiment}, the end effector follows the configuration "Motion~1" towards goal \((g_1)\) until the condition \({pos_{p,z} > p_1}\) of the branch "Motion~2" is fulfilled. This branch is ticked before and executes if the condition is fulfilled.

The parameters that need to be determined include both conditions of nodes \(p_1\) and \(p_2\) as well as the position of the goal points \(g_1\) and \(g_2\) along two axis.
The structure of the BT has redundancies since the thresholds and the positions of the goal points interplay with each other.

\begin{figure}[tpb]
	\centering
	{
		\setlength{\fboxrule}{0pt}
		\framebox{\parbox{3in}{
				\includegraphics[trim=5 8 0 36, clip, width=1.0\columnwidth]{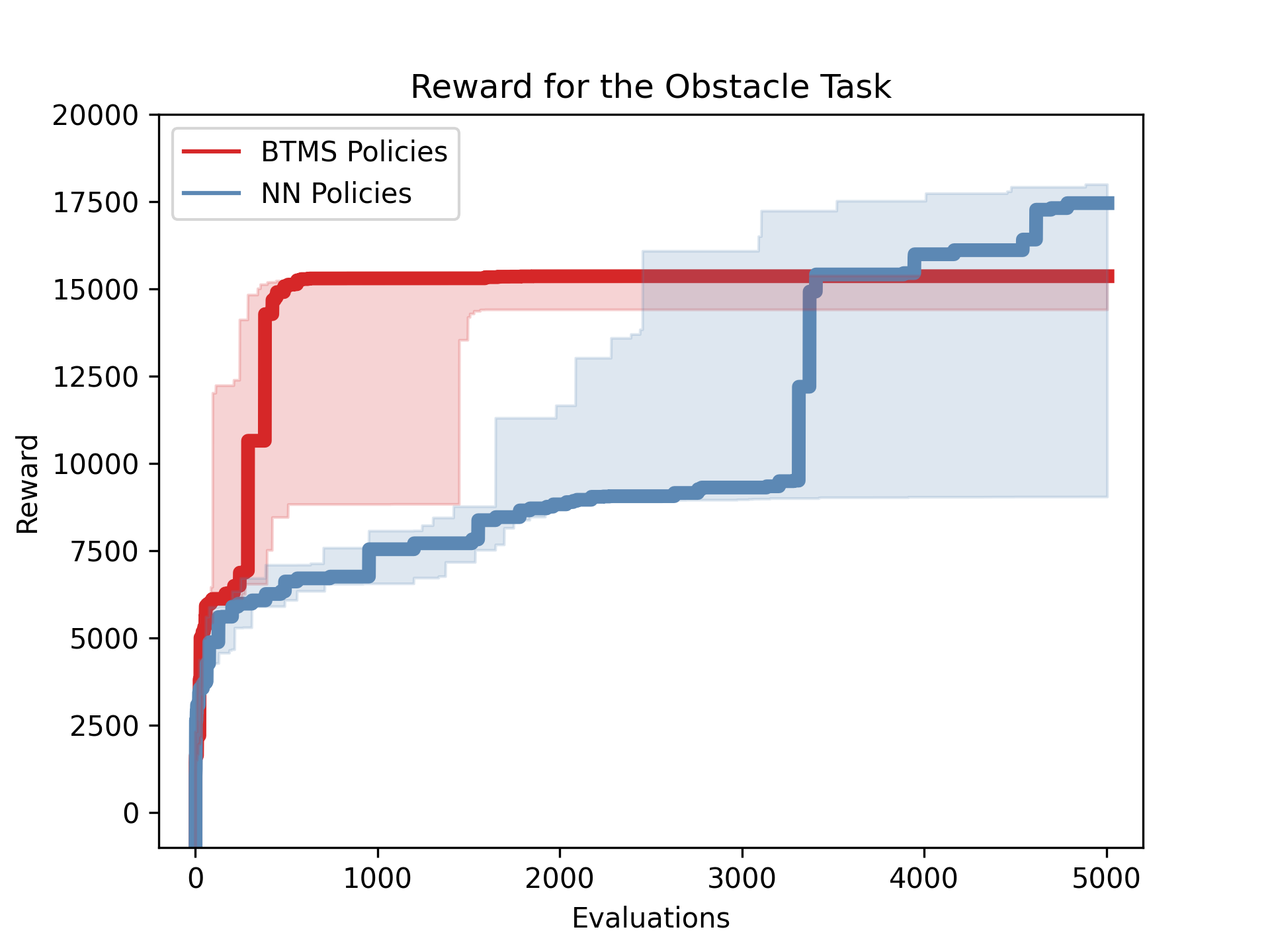}
				\caption{The development of the best reward when learning the obstacle task in simulation. The lines are the median values and the shaded region the 25\(^{th}\) and 75\(^{th}\) percentile (10 replicates). BT policies converge faster and with less variance.}
				\label{fig:obstacle-results}
	}}}

\end{figure}

We learned 10 policies with 5000 iterations each. 
All policies that successfully executed in simulation also successfully executed on the real system and there were no collisions with the environment.

We compare our approach to a feed-forward neural net that has one hidden layer with ten neurons. The function of the $i^{t h}$ layer is $\mathbf{y}_{i}=\phi_{i}\left(\mathbf{W}_{i} \mathbf{y}_{i-1}+\mathbf{b}_{i}\right),$ with $\mathbf{W}_{i}$ and $\mathbf{b}_{i}$ being the weight matrix and bias vector.
$\mathbf{y}_{i-1}$ and $\mathbf{y}_{i}$ are the input and output vectors and $\phi_{i}$ is the activation function. We use the hyperbolic tangent as the activation function for all the layers. The output layer sets the target position of the end effector.

The development of the reward in Fig.~\ref{fig:obstacle-results} shows that BTMS policies learned significantly faster than neural nets. Neural nets could eventually outperform the given structure of the movement skills. However, even though they control the end effector goal position, the motions can be much harder to predict. In fact, it is also not clear how this neural net-based policy will handle perturbations. In case of the BTMS policy, the underlying BT of the policy will simply evaluate the end-effector pose making thus the reaction predictable.

\subsection{Peg-in-Hole Task}

The peg-in-hole task is a common assembly task, e.g., it resembles the piston insertion from~\cite{rovida182iicirsi}.
The setup for the experiment is shown in Fig.~\ref{fig:experiment}. We know the position of the box within the frame of the workstation with a precision error of less than \SI{10}{\mm}. Based on the idea from~\cite{rovida182iicirsi} we allow the robot to make a search movement close to the surface of the box in order to find the hole. Intuitively, this is similar to the movement humans would do as well.

\begin{figure}[tpb]
	\centering
	{
		\setlength{\fboxrule}{0pt}
		\framebox{\parbox{3in}{
			\includegraphics[trim=5 8 0 38., clip, width=1.0\columnwidth]{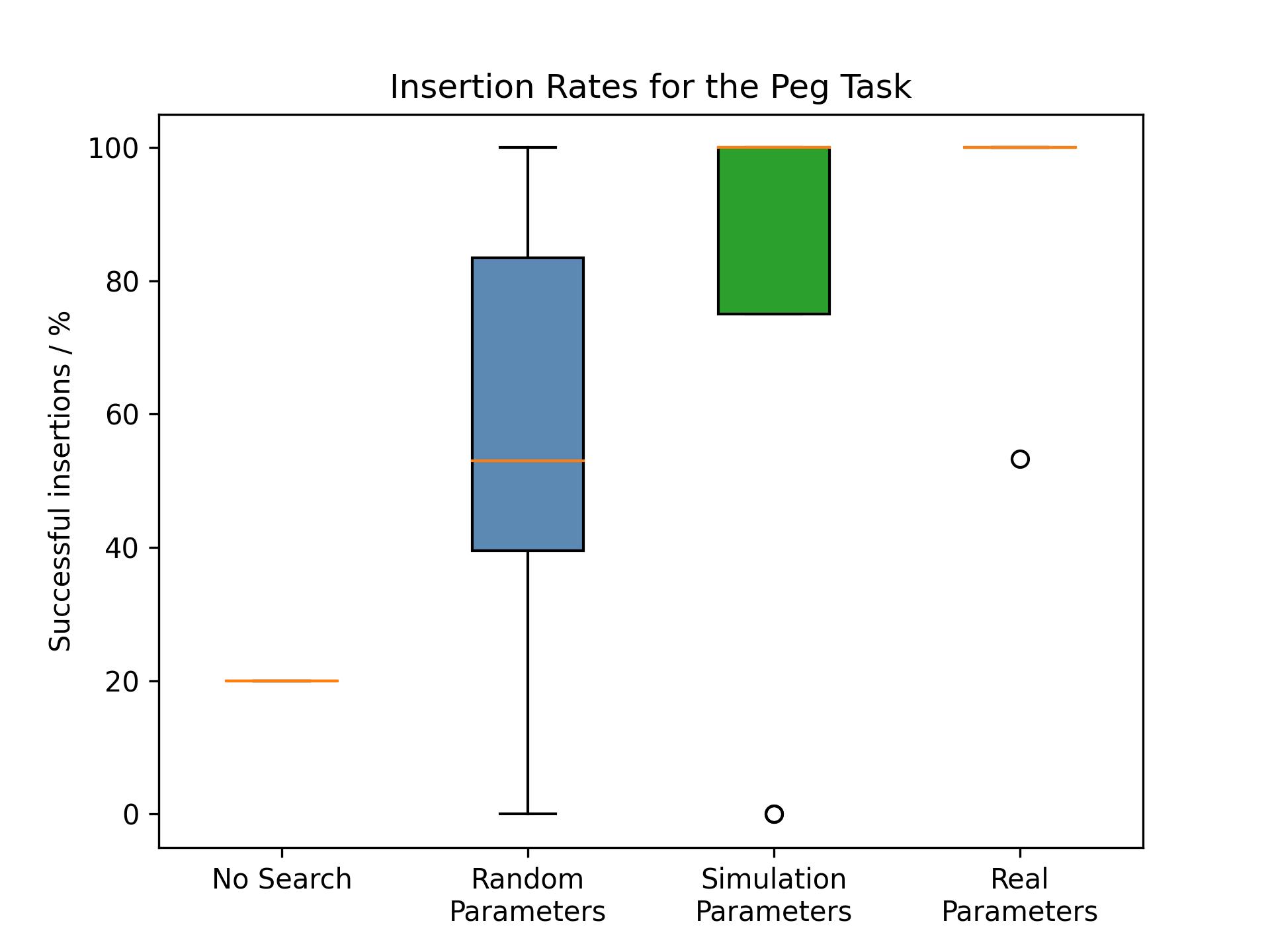}
			\caption{The success rates of the peg-in-hole experiment (15 trials per policy). The box plots show the median (red line) and interquartile range (25\(^{th}\) and 75\(^{th}\) percentile); the whiskers extend to the most extreme data points not considered outliers, and outliers are plotted individually. The learned search parameters clearly improve the insertion rate. Parameters learned on the physical robot have less negative outliers.}
			\label{fig:peg-results}
	}}}
\end{figure}

Tasks that involve contact forces are much harder to learn than free-space movements. We learn both the \(z\)-coordinate of the goal point \(g_3\) that influences the pressure that is applied on the surface and the search path velocity \(v_s\). The peg can get stuck or can jump over the hole if there is too little pressure for a given path velocity \(v_s\). The radius of the peg is \SI{2.5}{\mm} smaller than the hole, and an insertion is called successful if the peg is more than \SI{10}{\mm} inside the hole.

In our particular setup, the robot is configured in a low stiffness mode to ensure safety guarantees in human-robot collaborative workspaces. This also means that there can be deviations from the path and that it can be perturbed by a human operator. Therefore, we learn the task with five different start positions and expect to learn a policy that generalizes to other configurations.

We learn policies with 5000 iterations in simulation. The episode length is set to \SI{15}{\s} and each episode starts from one of the five random start positions and a random displacement of the hole. Again, learning in simulation allows to highly parallelize the procedure and does not block or damage expensive hardware.

The learned policies are evaluated on the physical setup with the five start positions used for training and ten previously unknown positions (15 in total). All start positions are \SI{40}{\cm} above the hole and have deviations in $x$ and $y$ direction between \SI{2}{\cm} and \SI{5}{\cm}. One position is exactly above the hole. We allow an episode length of \SI{25}{\s}.

The results are presented in Fig.~\ref{fig:peg-results}.
As a baseline we use a peg insertion without a search (Fig.~\ref{fig:peg-results}, left) and randomly sampled search parameters (Fig.~\ref{fig:peg-results}, second from left).
The results show that without a search motion a success rate of only \SI{20}{\percent} was achieved. 
Randomly sampled parameters had a median insertion rate of only \SI{53}{\percent}.
The parameters learned in simulation (Fig.~\ref{fig:peg-results}, third from left) performed with a median success rate of \SI{100}{\percent} generally well, but had negative outliers. However, these are easy to detect using the success criterion of the BT since they also perform significantly worse in simulation.

We acknowledge that an accurate enough simulation of a contact-rich task might not always be available. Therefore, we also learned six policies with 200 iterations on the real robot. With the episode length of \SI{15}{\s} and approximately \SI{6}{\s} to reset, the training took about \SI{75}{\min} per policy. The learning is safe in the sense that the robot can not collide with other objects in the workplace.
The parameters learned with the real system showed the most consistent performance with a median success rate of \SI{100}{\percent} and only one outlier that still achieved \SI{53}{\percent}.

\subsection{Combining the Tasks}

The modular nature of the BTs allows us to combine the policies of the two tasks into one or split a large policy into sub-policies. In this experiment we want to explore 1) if two policies that are learned in simulation and reality, respectively, can be combined, 2) how well the combined policies perform and 3) to which extend it can accomplish the task if the hole is displaced.

We combine the obstacle task policies learned in simulation with the peg insertion policies learned with the real system one-to-one and evaluated them on the BT shown in Fig.~\ref{fig:bt-task}. We assessed if they 1) can be executed on the robot, 2) collide with the environment and 3) successfully insert the peg. In addition to the hole being in the specified location, we also evaluated with deviations of \(\pm\)\SI{5}{\mm} and \(\pm\)\SI{10}{\mm} in both horizontal directions \(x\) and \(y\). We allow an episode length of \SI{30}{\s}.

None of the policies collide with the environment. In case of a perfect positioning of the hole an insertion rate of \SI{100}{\percent} was achieved. A misalignment of \(\pm\)\SI{5}{\mm} led to a success rate of \SI{91.66}{\percent}, while a displacement of \SI{10}{\mm} still allowed an insertion rate of \SI{83.3}{\percent}. These results showcase the modularity and efficacy of our BTMS method.

\section{Conclusions and Future Work}
\label{sec:conclusion}
Motivated by ~\cite{rovida182iicirsi} we introduced a pipeline to learn tasks with the BTMS policy representation that uses BTs and parametric movement skills. It is an interpretable, robust and modular policy for robot manipulators that learns faster than black-box policies, e.g., neural nets. 
We use a black-box optimizer to learn the paramters of the BTMS policy in simulation.
Domain randomization assured that policies are robust enough to cope with differences between the DT and the real robot.

We demonstrated the BTMS policy learning with a free-space movement and a peg-in-hole task. We also showed how these two can be easily combined into a single, more complex policy. Furthermore, we demonstrate that sub-policies learned in simulation can be combined  with counterparts learned with the real manipulator. For many tasks this policy representations offers a low dimensional search space that even allows to learn with physical robots.

To the best of our knowledge this is the first application of RL to the parameterization of behavior trees with manipulators. This opens up a new way to use and parameterize BTs. Furthermore, this representation allows to easily combine learning with reasoning by providing prior knowledge in terms of the structure or the parameters of the BT. An obvious limitation is that learning cannot be successful if the policy is not flexible enough to allow solutions for a particular task.

We plan to align the simulation more with reality using parameterized prior models~\cite{chatzilygeroudis182iicrai} to further improve the performance for more challenging assembly tasks. We are also planning to look into learning tasks that require dual-arm coordination. Finally, the approach is not limited to movement skills, and we plan to use it with other parameterizable skills such as grasping policies.




\section*{APPENDIX}
The source code and the submitted video are available at~\url{https://github.com/matthias-mayr/behavior-tree-policy-learning}.
%

\section*{ACKNOWLEDGMENT}
This work was partially supported by the Wallenberg AI, Autonomous Systems and Software Program (WASP) funded by Knut and Alice Wallenberg Foundation.
This research was also supported in part by affiliate members and other supporters of the Stanford DAWN project—Ant Financial, Facebook, Google, InfoSys, Teradata, NEC, and VMware.

We want to thank Francesco Rovida, Bjarne Großmann, Alexander Dürr, Björn Olofsson and Julian Salt for the interesting discussions and the constructive feedback.

”\copyright~2021 IEEE. Personal use of this material is permitted. Permission from IEEE must be obtained for all other uses, in any current or future media, including reprinting/republishing this material for advertising or promotional purposes, creating new collective works, for resale or redistribution to servers or lists, or reuse of any copyrighted component of this work in other works.”


\bibliography{root}{}

\begin{thebibliography}{10}
\providecommand{\url}[1]{#1}
\csname url@samestyle\endcsname
\providecommand{\newblock}{\relax}
\providecommand{\bibinfo}[2]{#2}
\providecommand{\BIBentrySTDinterwordspacing}{\spaceskip=0pt\relax}
\providecommand{\BIBentryALTinterwordstretchfactor}{4}
\providecommand{\BIBentryALTinterwordspacing}{\spaceskip=\fontdimen2\font plus
\BIBentryALTinterwordstretchfactor\fontdimen3\font minus
  \fontdimen4\font\relax}
\providecommand{\BIBforeignlanguage}[2]{{%
\expandafter\ifx\csname l@#1\endcsname\relax
\typeout{** WARNING: IEEEtran.bst: No hyphenation pattern has been}%
\typeout{** loaded for the language `#1'. Using the pattern for}%
\typeout{** the default language instead.}%
\else
\language=\csname l@#1\endcsname
\fi
#2}}
\providecommand{\BIBdecl}{\relax}
\BIBdecl

\bibitem{chatzilygeroudis2019survey}
K.~Chatzilygeroudis, V.~Vassiliades, F.~Stulp, S.~Calinon, and J.-B. Mouret,
  ``A survey on policy search algorithms for learning robot controllers in a
  handful of trials,'' \emph{IEEE Transactions on Robotics}, vol.~36, no.~2,
  pp. 328--347, 2019.

\bibitem{deisenroth13r}
\BIBentryALTinterwordspacing
M.~P. Deisenroth, G.~Neumann, and J.~Peters, ``A {{Survey}} on {{Policy
  Search}} for {{Robotics}},'' \emph{Foundations and Trends® in Robotics},
  vol.~2, no. 1–2, pp. 1--142, 2013. [Online]. Available:
  \url{https://www.nowpublishers.com/article/Details/ROB-021}
\BIBentrySTDinterwordspacing

\bibitem{khansari2011learning}
S.~M. Khansari-Zadeh and A.~Billard, ``Learning stable nonlinear dynamical
  systems with gaussian mixture models,'' \emph{IEEE Transactions on Robotics},
  vol.~27, no.~5, pp. 943--957, 2011.

\bibitem{ude2010task}
A.~Ude, A.~Gams, T.~Asfour, and J.~Morimoto, ``Task-specific generalization of
  discrete and periodic dynamic movement primitives,'' \emph{IEEE Transactions
  on Robotics}, vol.~26, no.~5, pp. 800--815, 2010.

\bibitem{stulp2013robot}
F.~Stulp and O.~Sigaud, ``Robot skill learning: From reinforcement learning to
  evolution strategies,'' \emph{Paladyn, Journal of Behavioral Robotics},
  vol.~4, no.~1, pp. 49--61, 2013.

\bibitem{ijspeert2013dynamical}
A.~J. Ijspeert, J.~Nakanishi, H.~Hoffmann, P.~Pastor, and S.~Schaal,
  ``Dynamical movement primitives: learning attractor models for motor
  behaviors,'' \emph{Neural computation}, vol.~25, no.~2, pp. 328--373, 2013.

\bibitem{krueger12icra}
V.~{Krüger}, V.~{Tikhanoff}, L.~{Natale}, and G.~{Sandini}, ``Imitation
  learning of non-linear point-to-point robot motions using dirichlet
  processes,'' in \emph{2012 IEEE International Conference on Robotics and
  Automation}, 2012, pp. 2029--2034.

\bibitem{rovida182iicirsi}
F.~Rovida, D.~Wuthier, B.~Grossmann, M.~Fumagalli, and V.~Krüger, ``Motion
  {{Generators Combined}} with {{Behavior Trees}}: {{A Novel Approach}} to
  {{Skill Modelling}},'' in \emph{2018 {{IEEE}}/{{RSJ International
  Conference}} on {{Intelligent Robots}} and {{Systems}} ({{IROS}})}, 2018, pp.
  5964--5971.

\bibitem{colledanchise142iicirsa}
M.~Colledanchise and P.~Ögren, ``How {{Behavior Trees}} modularize robustness
  and safety in hybrid systems,'' in \emph{2014 {{IEEE}}/{{RSJ International
  Conference}} on {{Intelligent Robots}} and {{Systems}}}, 2014, pp.
  1482--1488.

\bibitem{colledanchise17ac}
------, \emph{Behavior {{Trees}} in {{Robotics}} and {{AI}}: {{An
  Introduction}}}.\hskip 1em plus 0.5em minus 0.4em\relax Chapman \& Hall/CRC
  Press, 2017.

\bibitem{chatzilygeroudis172iicirsi}
K.~Chatzilygeroudis, R.~Rama, R.~Kaushik, D.~Goepp, V.~Vassiliades, and J.-B.
  Mouret, ``Black-box data-efficient policy search for robotics,'' in
  \emph{2017 {{IEEE}}/{{RSJ International Conference}} on {{Intelligent
  Robots}} and {{Systems}} ({{IROS}})}, 2017, pp. 51--58.

\bibitem{tobin17iros}
J.~{Tobin}, R.~{Fong}, A.~{Ray}, J.~{Schneider}, W.~{Zaremba}, and P.~{Abbeel},
  ``Domain randomization for transferring deep neural networks from simulation
  to the real world,'' in \emph{2017 IEEE/RSJ International Conference on
  Intelligent Robots and Systems (IROS)}, 2017, pp. 23--30.

\bibitem{mehta20pmlr}
B.~Mehta, M.~Diaz, F.~Golemo, C.~J. Pal, and L.~Paull, ``Active domain
  randomization,'' in \emph{Proceedings of the Conference on Robot Learning},
  ser. Proceedings of Machine Learning Research, L.~P. Kaelbling, D.~Kragic,
  and K.~Sugiura, Eds., vol. 100, 2020, pp. 1162--1176.

\bibitem{rovida172iicirsi}
F.~Rovida, B.~Grossmann, and V.~Krüger, ``Extended behavior trees for quick
  definition of flexible robotic tasks,'' in \emph{2017 {{IEEE}}/{{RSJ
  International Conference}} on {{Intelligent Robots}} and {{Systems}}
  ({{IROS}})}, 2017, pp. 6793--6800.

\bibitem{polydoros17jirs}
\BIBentryALTinterwordspacing
A.~S. Polydoros and L.~Nalpantidis, ``Survey of {{Model}}-{{Based Reinforcement
  Learning}}: {{Applications}} on {{Robotics}},'' \emph{Journal of Intelligent
  \& Robotic Systems}, vol.~86, no.~2, pp. 153--173, 2017. [Online]. Available:
  \url{https://doi.org/10.1007/s10846-017-0468-y}
\BIBentrySTDinterwordspacing

\bibitem{iovino2020survey}
M.~Iovino, E.~Scukins, J.~Styrud, P.~Ögren, and C.~Smith, ``A survey of
  behavior trees in robotics and ai,'' 2020.

\bibitem{marzinotto142iicrai}
A.~Marzinotto, M.~Colledanchise, C.~Smith, and P.~Ögren, ``Towards a unified
  behavior trees framework for robot control,'' in \emph{2014 {{IEEE
  International Conference}} on {{Robotics}} and {{Automation}} ({{ICRA}})},
  2014, pp. 5420--5427.

\bibitem{7989070}
C.~{Paxton}, A.~{Hundt}, F.~{Jonathan}, K.~{Guerin}, and G.~D. {Hager},
  ``Costar: Instructing collaborative robots with behavior trees and vision,''
  in \emph{2017 IEEE International Conference on Robotics and Automation
  (ICRA)}, 2017, pp. 564--571.

\bibitem{lim10aec}
C.-U. Lim, R.~Baumgarten, and S.~Colton, ``Evolving {{Behaviour Trees}} for the
  {{Commercial Game DEFCON}},'' in \emph{Applications of {{Evolutionary
  Computation}}}, ser. Lecture {{Notes}} in {{Computer Science}}, C.~Di~Chio,
  S.~Cagnoni, C.~Cotta, M.~Ebner, A.~Ekárt, A.~I. Esparcia-Alcazar, C.-K. Goh,
  J.~J. Merelo, F.~Neri, M.~Preuß, J.~Togelius, and G.~N. Yannakakis,
  Eds.\hskip 1em plus 0.5em minus 0.4em\relax {Springer}, 2010, pp. 100--110.

\bibitem{perez11aeca}
D.~Perez, M.~Nicolau, M.~O’Neill, and A.~Brabazon, ``Evolving {{Behaviour
  Trees}} for the {{Mario AI Competition Using Grammatical Evolution}},'' in
  \emph{Applications of {{Evolutionary Computation}}}, ser. Lecture {{Notes}}
  in {{Computer Science}}, C.~Di~Chio, S.~Cagnoni, C.~Cotta, M.~Ebner,
  A.~Ekárt, A.~I. Esparcia-Alcázar, J.~J. Merelo, F.~Neri, M.~Preuss,
  H.~Richter, J.~Togelius, and G.~N. Yannakakis, Eds.\hskip 1em plus 0.5em
  minus 0.4em\relax {Springer Berlin Heidelberg}, 2011, pp. 123--132.

\bibitem{colledanchise18itg}
M.~Colledanchise, R.~N. Parasuraman, and P.~Ogren, ``Learning of {{Behavior
  Trees}} for {{Autonomous Agents}},'' \emph{IEEE Transactions on Games}, pp.
  1--1, 2018.

\bibitem{fu16a}
\BIBentryALTinterwordspacing
Y.~Fu, L.~Qin, and Q.~Yin, ``A {{Reinforcement Learning Behavior Tree
  Framework}} for {{Game AI}},'' in \emph{{International Conference on
  Economics, Social Science, Arts, Education and Management
  Engineering}}.\hskip 1em plus 0.5em minus 0.4em\relax {Atlantis Press}, 2016.
  [Online]. Available:
  \url{https://www.atlantis-press.com/proceedings/essaeme-16/25860289}
\BIBentrySTDinterwordspacing

\bibitem{park17itie}
H.~{Park}, J.~{Park}, D.~{Lee}, J.~{Park}, M.~{Baeg}, and J.~{Bae},
  ``Compliance-based robotic peg-in-hole assembly strategy without force
  feedback,'' \emph{IEEE Transactions on Industrial Electronics}, vol.~64,
  no.~8, pp. 6299--6309, 2017.

\bibitem{topp182iicirsi}
E.~A. Topp, M.~Stenmark, A.~Ganslandt, A.~Svensson, M.~Haage, and J.~Malec,
  ``Ontology-{{Based Knowledge Representation}} for {{Increased Skill
  Reusability}} in {{Industrial Robots}},'' in \emph{2018 {{IEEE}}/{{RSJ
  International Conference}} on {{Intelligent Robots}} and {{Systems}}
  ({{IROS}})}, 2018, pp. 5672--5678.

\bibitem{hansen2006towards}
N.~Hansen, ``Towards a new evolutionary computation,'' \emph{Studies in
  Fuzziness and Soft Computing}, vol. 192, pp. 75--102, 2006.

\bibitem{hansen2009benchmarking}
------, ``Benchmarking a {{BI}}-population {{CMA}}-{{ES}} on the {{BBOB}}-2009
  function testbed,'' in \emph{Proceedings of the 11th Annual Conference
  Companion on Genetic and Evolutionary Computation Conference: {{Late}}
  Breaking Papers}, 2009, pp. 2389--2396.

\bibitem{cully18j}
A.~Cully, K.~Chatzilygeroudis, F.~Allocati, and J.-B. Mouret, ``Limbo: {{A
  Flexible High}}-performance {{Library}} for {{Gaussian Processes}} modeling
  and {{Data}}-{{Efficient Optimization}},'' \emph{Journal of Open Source
  Software}, vol.~3, no.~26, p. 545, 2018.

\bibitem{nguyen18iros}
P.~D.~H. {Nguyen}, T.~{Fischer}, H.~J. {Chang}, U.~{Pattacini}, G.~{Metta}, and
  Y.~{Demiris}, ``Transferring visuomotor learning from simulation to the real
  world for robotics manipulation tasks,'' in \emph{2018 IEEE/RSJ International
  Conference on Intelligent Robots and Systems (IROS)}, 2018, pp. 6667--6674.

\bibitem{chen18iros}
X.~{Chen}, A.~{Ghadirzadeh}, J.~{Folkesson}, M.~{Björkman}, and P.~{Jensfelt},
  ``Deep reinforcement learning to acquire navigation skills for wheel-legged
  robots in complex environments,'' in \emph{2018 IEEE/RSJ International
  Conference on Intelligent Robots and Systems (IROS)}, 2018, pp. 3110--3116.

\bibitem{koos2012transferability}
S.~Koos, J.-B. Mouret, and S.~Doncieux, ``The transferability approach:
  Crossing the reality gap in evolutionary robotics,'' \emph{IEEE Transactions
  on Evolutionary Computation}, vol.~17, no.~1, pp. 122--145, 2012.

\bibitem{lee18joss}
\BIBentryALTinterwordspacing
J.~Lee, M.~Grey, S.~Ha, T.~Kunz, S.~Jain, Y.~Ye, S.~Srinivasa, M.~Stilman, and
  C.~Liu, ``{{DART}}: {{Dynamic Animation}} and {{Robotics Toolkit}},''
  \emph{Journal of Open Source Software}, vol.~3, no.~22, p. 500, 2018.
  [Online]. Available: \url{https://joss.theoj.org/papers/10.21105/joss.00500}
\BIBentrySTDinterwordspacing

\bibitem{deisenroth11p2icicml}
M.~P. Deisenroth and C.~E. Rasmussen, ``{{PILCO}}: A model-based and
  data-efficient approach to policy search,'' in \emph{Proceedings of the 28th
  {{International Conference}} on {{International Conference}} on {{Machine
  Learning}}}, ser. {{ICML}}'11.\hskip 1em plus 0.5em minus 0.4em\relax
  {Omnipress}, 2011, pp. 465--472.

\bibitem{chatzilygeroudis182iicrai}
K.~Chatzilygeroudis and J.-B. Mouret, ``Using {{Parameterized Black}}-{{Box
  Priors}} to {{Scale Up Model}}-{{Based Policy Search}} for {{Robotics}},'' in
  \emph{2018 {{IEEE International Conference}} on {{Robotics}} and
  {{Automation}} ({{ICRA}})}, 2018, pp. 5121--5128.

\end{thebibliography}
\bibliographystyle{IEEEtran}

\end{document}